\documentclass[runningheads]{llncs}

\usepackage{tikz}
\usepackage{url}
\usepackage{multirow}
\usepackage[utf8]{inputenc}
\usepackage[T1]{fontenc}

\begin{document}

\title{Taking Class Imbalance Into Account in Open~Set~Recognition Evaluation}

\author{Joanna~Komorniczak~\orcidID{0000-0002-1393-3622} \and
Paweł~Ksieniewicz~\orcidID{0000-0001-9578-8395}}
\authorrunning{J. Komorniczak and P. Ksieniewicz}
%
\institute{Department of Systems and Computer Networks, \\
Wrocław University of Science and Technology,\\
Wyb. Wyspianskiego 27, 50370 Wrocław, Poland\\
\email{\{joanna.komorniczak,pawel.ksieniewicz\}@pwr.edu.pl}}

\maketitle

\begin{abstract}
In recent years Deep Neural Network-based systems are not only increasing in popularity but also receive growing user trust. However, due to the \textit{closed-world assumption} of such systems, they cannot recognize samples from unknown classes and often induce an incorrect label with high confidence. Presented work looks at the evaluation of methods for Open Set Recognition, focusing on the impact of class imbalance, especially in the dichotomy between known and unknown samples. As an outcome of problem analysis, we present a set of guidelines for evaluation of methods in this field.
\end{abstract}

\keywords{Open Set Recognition \and Classification \and Neural Networks \and Imbalanced Data Classification}

\section{Introduction}

Machine learning algorithms and models are increasingly used to assist people in making everyday decisions~\cite{voulodimos2018deep}. What is more, a contemporary user, due to the spreading popularity of \textsc{ai}-aided solutions, increasingly trusts automatic recognition systems. The universality and high recognition quality of such solutions result from vast computational resources~\cite{liu2017survey}, allowing for excessively long-term training to bring the quality of recognition of the validation set close to the limit of perfection. However, preparing a model for all unknown inputs in the training process, including objects of new classes, is still impossible~\cite{masana2022class}.

Meanwhile, most machine learning models will not be able to recognize that the analyzed object does not belong to the distribution of a problem represented by its training data and, therefore, will, with high certainty, assign the unknown object to one of the classes used in the training process \cite{geng2020recent}. Moreover, the explainability of such solutions is still an open research topic~\cite{gunning2019xai}, showing vulnerabilities to threats like \textit{Adversarial attacks} \cite{choras2021double}, in which minor manipulations of input data, imperceptible to a human, can fool \textit{Deep Learning} algorithms and lead to a false answer~\cite{rosenberg2021adversarial}. 

Modern recognition systems adapted to the real needs of users, and not just to standard experimental protocols~\cite{li2020model}, should be able to both effectively distinguish objects between closed sets of classes and recognize when the input data in the inference procedure has been manipulated or represents instances that lie far outside the sampling area of the training set~\cite{boult2019learning}. Such a specific environment makes the experimental evaluation of artificial intelligence models in terms of effectiveness in Open Set tasks a severe challenge, requiring very high diligence and the use of possibly sensitive and broad analytical framework. 

\subsection{Open Set Recognition}

Classic classification methods were designed to carry out recognition within a finite space and a finite set of classes~\cite{bellman1966abstraction}. The mapping of the entire problem space to a closed set of classes is, by default, a fully solved task, i.e., one in which every query will result in a decision. Such an approach resembles laboratory tests, which are extremely difficult to reproduce in an environment of an ever-changing Open World when the recognized categories evolve after production phase of a system~\cite{kolajo2019big}, extraction models are adapted to new problems through \textit{Transfer Learning}~\cite{zhu2020transfer}, definitions of known problems do extend with new classes, and thus new, unknown threats for the reliability of prediction emerge.

Open Set Recognition (\textsc{osr}) methods are designed to perform pattern recognition in potentially infinite space. A proper solution of this type combines two machine learning tasks -- \textit{novelty detection} and \textit{multiclass classifiaction}, so the task of the \textsc{osr} algorithm is both to recognize within known classes and to respond to unknown objects with possibly minimized trade-off. While a routine classifier aims to minimize the risk of making an incorrect decision \cite{wozniak2013hybrid}, the purpose of classifiers in Open Set is to balance the risk of making an incorrect decision (\textit{empirical risk}) and the risk of recognizing an unknown object as an instance of a known class (\textit{risk of the unknown})~\cite{scheirer2012toward}.

\emph{Risk of the unknown} is associated with an unbounded open space, typical for a large pool of currently used classification algorithms such as neural networks with a \textit{Softmax} layer, linear classifiers, and other traditional models such as {$k$-Nearest Neighbors} and \emph{Gaussian Na\"ive Bayes}. Unbounded open space becomes an issue when the classifier assigns an infinite area to one of the known classes. When instances of an unknown class appear -- often located far from the decision boundary -- such algorithms will assign unknown instances with high confidence to one of the classes used to train the model \cite{boult2019learning}. \emph{Open Set Recognition} algorithms are designed to limit the scope of the classifier's competence only to the area sampled by the available training set \cite{bendale2016towards}. Thus, an instance lying \textit{far} from known samples should be recognized as an instance of the unknown class -- identified with no available records in the training set. Unfortunately, such an approach often loses its intuition in deep neural networks due to the observed \emph{feature collapse} phenomenon, which forces objects from initially significantly distant classes to transform into a similar region of a final embedding space during the deep feature extraction~\cite{van2020uncertainty}. 

\subsection{Class taxonomy}

In the currently recognized taxonomy of classes, the \textsc{osr} task operates on two basic categories of classes: 
\begin{itemize}
    \item \textit{Known Known Classes} (\textsc{kkc}), whose instances were used to train the model,
    \item \textit{Unknown Unknown Classes} (\textsc{uuc}), which are considered instances of unknown classes, used only in the testing process \cite{geng2020recent}.
\end{itemize}

The other two class categories -- \textit{Known Unknown Classes} (\textsc{kuc}) and \textit{Unknown Known Classes} (\textsc{ukc}) -- are typical for detection tasks and Zero-shot learning \cite{pourpanah2022review}. \textsc{kuc} are classes whose instances are present in the available sampling, and there is initial knowledge that these classes are not recognized in the learning task. A proper example might be the recognition of digits when the algorithm is additionally fed with examples of letters with the assumption of letters not being an object of interest. \textsc{ukc} are classes for which instances are not observed, but some information describing the characteristics of the data is available.

\subsection{Contribution and motivation}

The main goal of this work is to emphasize common difficulties in evaluating Open Set Recognition methods, coming primarily from the imbalanced nature of the task they are solving. The conducted research will critically analyze the techniques employed for evaluating \textsc{osr} methods, concluding with a list of recommendations for correct validation in the presence of varying \textsc{kkc}/\textsc{uuc} class distribution. 

The work presents an exemplary model quality assessment to motivate the proposed extension of the experimental protocol. It is performed using four measures based on Balanced Accuracy Score -- two commonly used in the literature for closed set classification (\textit{Inner score}) and open set detection (\textit{Outer score})~\cite{neal2018open}, and two novel approaches (\textit{Halfpoint} and \textit{Overall} scores) -- adapted to the specificity of Open Set Recognition.

The main contributions of the presented work are:
\begin{itemize}
     \item The analysis of common techniques for evaluating the quality of methods' operation, considering the importance of class imbalance, experimental protocol, and metric selection.
     \item An extension of measures used to access Open Set Recognition quality to the total of four measures -- \textit{Inner}, \textit{Outer}, \textit{Halfpoint}, and \textit{Overall} scores.
\end{itemize}

\section{Related works}

The early articles in the field of \emph{Open Set Recognition} employed classical machine learning methods unrelated to deep learning. The first algorithm to address the \textsc{osr} problem was modifying the Linear Support Vector Machine (SVM)~\cite{scheirer2012toward}. However, this method did not formally solve \textsc{osr} by not bounding Open Space Risk \cite{boult2019learning}, which means that there was a possibility that unknown samples \textit{far} from the training instances will be recognized as \textsc{kkc}. Later proposed methods, based on SVM with a non-linear kernel, used the Compact Abating Probability (CAP) model~\cite{scheirer2014probability}, formally solved \textsc{osr} problems by limiting the recognition area of \textsc{kkc} instances. Those methods employed Extreme Value Theory in Weibull-calibrated SVM \cite{scheirer2014probability} and $P_I$-SVM algorithms \cite{jain2014multi}, applying the threshold-based classification scheme. 

Another family of proposed methods were distance-based solutions, such as Nearest Non-Outlier (NNO) \cite{bendale2015towards}, which takes into account the affinity with \textsc{kkc}, and Open Set Nearest Neighbor (OSNN)~\cite{mendes2017nearest}, which makes decisions based on the ratio of the distance between the two nearest neighbors.

Addressing unknown inputs in methods based on deep neural networks is particularly important due to unsupervised feature extraction \cite{lecun2015deep,boult2019learning}, limited explainability of deep learning methods \cite{marcinkevivcs2020interpretability} and \emph{feature collapse} phenomenon~\cite{van2020uncertainty}. The Softmax layer, dedicated to closed-set classification, has become a standard in neural architectures, which, not adapted to recognition in an open set, will recognize unknown instances as \textsc{kkc} objects with high certainty. The first method directly related to neural networks was \textit{Openmax}~\cite{bendale2016towards}, using Weibull distributions, adjusted to the distance of the training instances from the Mean Activation Vector (MAV) of their classes. Based on the fitted distributions, the pseudo-activations are generated for the class describing unknown objects. Openmax also uses a mechanism for marking \textit{uncertain} objects that are close to the decision boundary and have low decision support. 

The method constituting the baseline for \textsc{osr} algorithms is a modified \emph{Softmax} (later denoted in this work as \emph{Tresholded Softmax}) operating analogously with a threshold defining the probability from which instances will be recognized as \textsc{kkc}. Uncertain objects below this threshold will be recognized as \textsc{uuc}. Despite the relative simplicity of this method, it has shown the ability to recognize unknown objects. The work by Pearce et al. \cite{pearce2021understanding} showed that objects not present in the training process can be mapped to the uncertain region, where, after thresholding, they can be recognized and marked. Other solutions replaced the last layer in networks with a 1-vs-set layer presenting the Deep Open Classifier (DOC)~\cite{shu2017doc}. In the Reciprocal Point Learning (RPL) \cite{chen2020learning} method, the authors use integrated points representing the unknown space to limit the area of the classifier's competence.

Especially in computer vision, generative methods have become quite common due to the promising results of similar approaches in the overall subject of data augmentation~\cite{antoniou2017data}. Based on the training data, the alleged \textsc{uuc} samples are synthetically generated in the learning process and later used in model training. One of the first methods of this type designed for \textsc{osr} was Generative Openmax (G-Openmax)~\cite{ge2017generative}, which extended Openmax by training the network with synthetically generated objects of an unknown class. Adversarial Sample Generation (ASG)~\cite{yu2017open}, in addition to generating \textsc{uuc} instances, can augment underrepresented \textsc{kkc}. Based on instances of known classes, the instances of \textsc{uuc} are also generated in the OSRCI~\cite{neal2018open} method. 

A generative model combined with a new neural network architecture proposal was presented in CROSR \cite{yoshihashi2019classification}. The Conditional Probabilistic Generative Models (CPGM)~\cite{sun2020open} framework, on the other hand, adds discriminant information to probabilistic generative models, allowing recognition within available classes and detection of unknown samples.

Per the current literature, all the generative models do not formally solve the \textsc{osr} problem, leaving \textit{unbounded open space risk}. Synthetic instances are usually generated from known classes, so they can occupy the \textit{uncertain} area rather than the \textit{unknown} space~\cite{boult2019learning}. In addition, \textsc{uuc} instances in generative models are modeled using instances of known classes, which results in intermediate instances specific to a given recognition problem. Nevertheless, the simplicity of using the generative framework, which may not even require any additional mechanisms at the network architecture level, and their high quality of recognition -- even in a clash with adversarial attacks -- make methods in this category often proposed and widely used \cite{xia2021adversarial}.

\section{Class Imbalance in Open Set Recognition}

The Open Set Recognition task consists of multiclass classification and recognition of unknown objects \cite{scheirer2012toward}. This results in a vast pool of alternatives in metrics for assessing the quality of method's operation. Their performance is usually evaluated using two components -- one assessing the model's ability to recognize objects in closed-set classification (denoted as \textit{Inner score}), the other responsible for the ability to recognize unknown classes (denoted as \textit{Outer} score). Most publications use the measures F-score \cite{scheirer2012toward,scheirer2014probability,jain2014multi}, AUC \cite{neal2018open,yoshihashi2019classification} and Accuracy \cite{ge2017generative,mendes2017nearest}.

\subsection{Common evaluation strategies}

Already in the first works in the field, the authors drew attention to the importance of imbalance in the \textsc{osr} problem \cite{scheirer2012toward}. The positive class, representing \textsc{kkc} samples, was supposed to be a less numerous set compared to the significant number of \textsc{uuc} instances in the infinite Open Space. As the authors state, the classic measure of accuracy may not correctly determine the quality of the tested model in the presence of a significant imbalance in the \textsc{kkc}/\textsc{uuc} dichotomy. Therefore, it was proposed to use the F-score measure, integrating the Recall and Precision. What is more, the available sources emphasize that the hallmark of correct evaluation is to perform experiments on sets described by different \textit{Openness}, depending on the number of \textsc{kkc} and \textsc{uuc} classes. 

The \textit{Openness} of the problem is an abstract concept developed solely for experimental evaluation. It depends on the number of classes used to train and test the model, and, therefore, on the number of \textsc{kkc} and \textsc{uuc}. The measure is calculated according to the formula:
\begin{equation}
        O = 1-\sqrt{\frac{2 \cdot |C_{TR}|}{|C_{TR}| + |C_{TE}|}},
\end{equation}
\noindent where $|C_{TR}|$ and $|C_{TE}|$ denote, respectively, the number of classes used for training, equivalent to the number of \textsc{kkc}, and the number of classes used to test the quality of the model, equivalent to the total number of \textsc{kkc} and \textsc{uuc} combined. 

The more classes belong to the \textsc{uuc} compared to \textsc{kkc}, the greater the \textit{Openness} of the problem. It is worth mentioning here, that in end-use applications, the only known feature of the problem is the number of available classes. Hence, the default values of hyperparameters should be determined according to this characteristic and not the \textit{Openness} itself. This measure does not consider the number of instances in the given classes, so it is possible that even with a high \textit{Openness} of the dataset, \textsc{uuc} samples will form a minority class. 

One of the most common experimental protocols uses \textsc{kkc} from the MNIST dataset and \textsc{uuc} from the Omniglot set during the evaluation \cite{yoshihashi2019classification}. A large number of classes available in the Omniglot dataset (above 900) allows evaluation of methods in the environment with significant \textit{Openness}. However, each class in the MNIST collection contains 7,000 objects, while the classes in the Omniglot collection contain 20 objects each. Such a disproportion between class sizes may result in significantly more known instances than unknown ones. The described configuration of the test set will lead to a high imbalance ratio, where \textsc{uuc}, marked as negatives \cite{scheirer2012toward}, form a minority class. In such a case, the F1-score metric may not reliably assess the recognition quality.

\begin{figure}[!htb]
    \centering
    \includegraphics[width=0.86\textwidth]{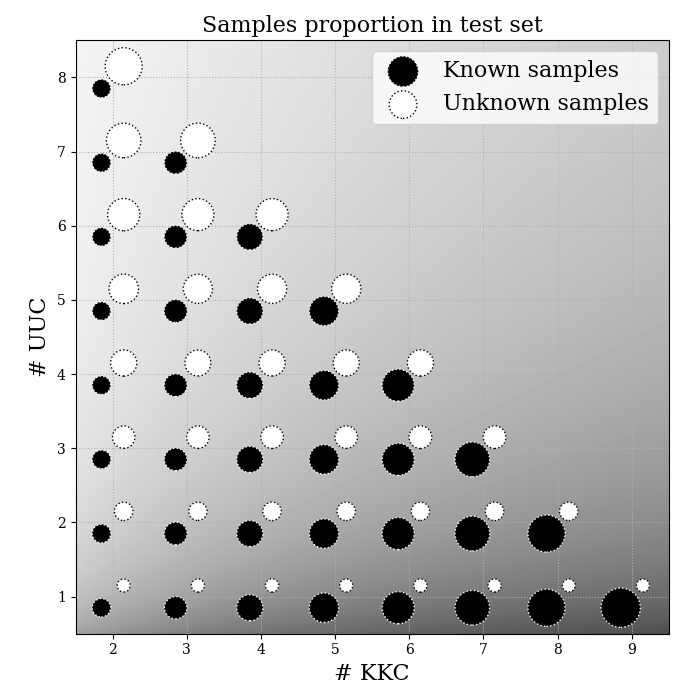}
    \caption{Number of samples coming from \textsc{kkc} and \textsc{uuc} in test set in relation to the dataset \textit{Openness} (function of \textsc{kkc} and \textsc{uuc} number)}
    \label{fig:owady_imb}
\end{figure}

Figure~\ref{fig:owady_imb} shows how the proportions between the number of instances of \textsc{kkc} and \textsc{uuc} classes in the test set change depending on the \textit{Openness} of the data set -- the number of \textsc{kkc} and \textsc{uuc}. The number of \textsc{kkc} is shown on the horizontal axis and \textsc{uuc} on the vertical axis. All possible configurations of the 10-class MNIST set were examined. The size of a black dot indicates the number of objects from the \textsc{kkc} in the test set, and the size of a white dot shows the number of objects from the \textsc{uuc}. It can be seen that the configuration of the problem (\textit{Openness}) affects the imbalance ratio of the binary problem of recognition between the \textsc{kkc} and \textsc{uuc} even if each class originally had an equal number of samples.

What is more, assuming the variability of the sets used to test \textsc{osr} models, it should be remembered that the F-score compares the precision of algorithms at different recalls \cite{dhamija2022five}, which may not give a correct estimation of recognition quality, especially in the case of monitoring algorithm's performance.

\subsection{Base metric impact on measured quality}

\begin{figure}[!htb]
    \centering
    \includegraphics[width=\textwidth]{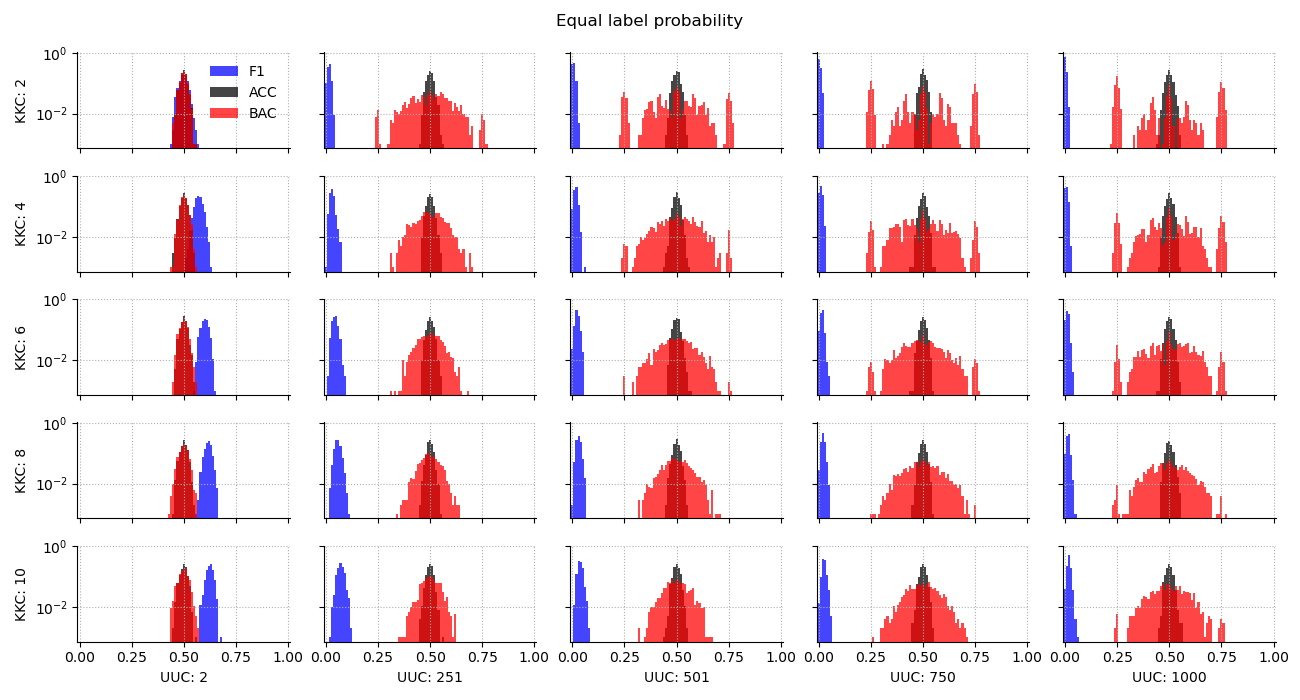}
    \includegraphics[width=\textwidth]{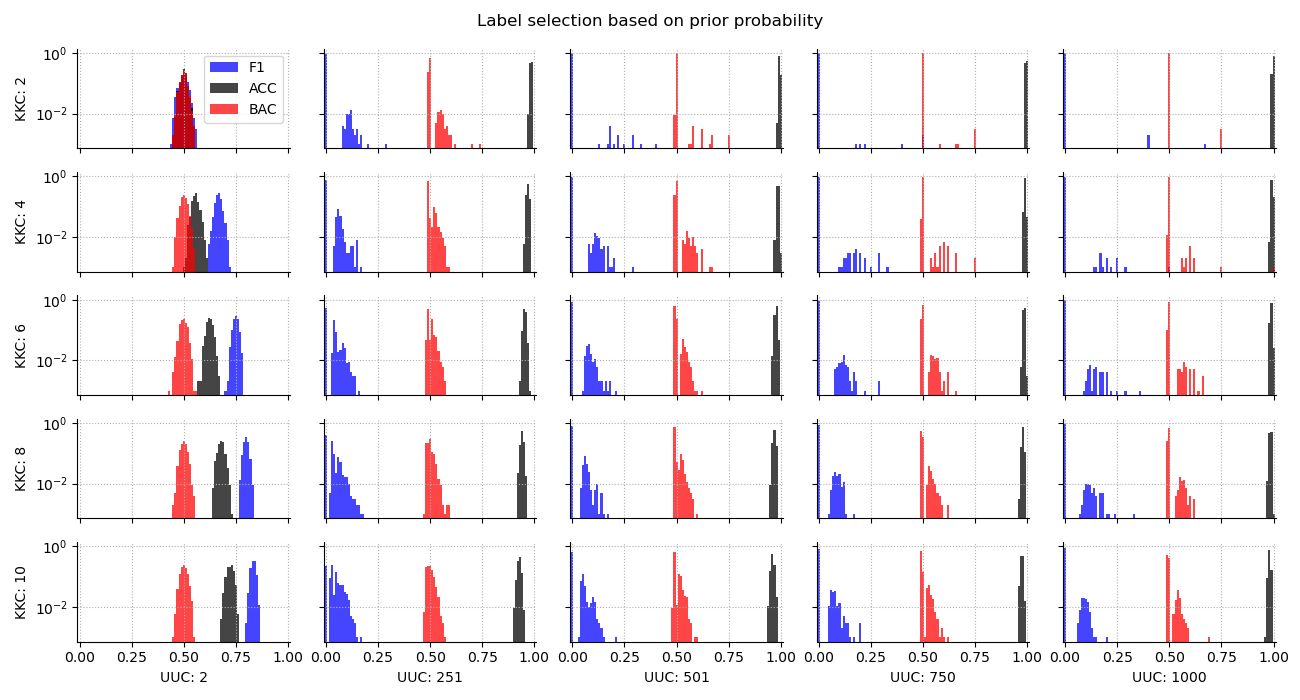}
    \caption{Histograms of measure values of random predictions for different configurations of \textsc{kkc}/\textsc{uuc} class numbers}
    \label{fig:buka}
\end{figure}

Figure~\ref{fig:buka} shows the \textit{Outer score} results of a simple experiment showing how the results are distributed in individual base metrics \cite{brzezinski2019dynamics} (F1, Accuracy and Balanced Accuracy were considered) in the case of random predictions. Each subplot presents a sample of one set used to evaluate \textsc{osr}. The \textsc{kkc} number is determined in the rows, and the \textsc{uuc} number on the columns. For example: the first subplot describes the case where two known and two unknown classes are used in the evaluation. For simplicity, it is assumed that the classes are of equal number, as in the case of the \textit{Holdout} protocol, sampling \textsc{kkc} and \textsc{uuc} from a common set. For the first subplot, the problem understood as the \textsc{kkc}/\textsc{uuc} dichotomy will be balanced. In the case of other configurations, we will observe a class imbalance. The most extreme cases are described by two known and a thousand unknown classes.

The charts present histograms of metric values after a thousand repetitions of random assignments of objects to classes. The first five rows of the figure show the results for the case when the classifier's random response proportions are default -- that is, it is equally likely to draw label 0 and 1. The next five rows show the results in case when the classifier is aware of the proportions within the problem -- when the probability of drawing a label will depend on actual proportions of the ground truth. It is worth emphasizing that in the case of \textsc{osr}, the number of \textsc{uuc} objects in relation to the number of \textsc{kkc} objects should remain unknown, but taking into account the case of extreme responses from the classifier should also be considered -- this is the purpose of presenting results dependent on prior probability.

It is easy to notice that the commonly used F1 measure (marked in blue) gets different values depending on the imbalance ratio within the \textsc{kkc}/\textsc{uuc} dichotomy. It is especially worth paying attention to the first column, where the \textsc{kkc} constitute the majority class, and the F1 metric, even for random predictions, achieves results of over 80\%. Accuracy, on the other hand, regardless of whether \textsc{kkc} or \textsc{uuc} is a minority class, will not reliably describe the results due to the accuracy paradox \cite{uddin2019addressing}. However, the expected value of the Balanced Accuracy Score measure remains around 50\%, which may indicate its potential in describing quality of recognition in this type of imbalance.
 
\subsection{Proposed experimental protocol extension}

This publication notes the significant importance of the number of \textsc{kkc} and \textsc{uuc} instances affecting the recognition problem's imbalance ratio. With many instances coming from \textsc{uuc} in the test set, we face common quality assessment problems in imbalanced tasks \cite{barros2019predictive}. 

The classic strategy of evaluating \textsc{osr} method based on the \textit{Inner score} (closed-set classification quality) seems highly debatable, where the protocol adopted by the literature does not take into account the prediction towards the unknown class and does not penalize the model for incorrectly identifying the object as unknown, giving it in a way \textit{second chance} and rewarding if \textit{second best shot} indicates the correct \textsc{kkc} category.

Reporting such a result only from the \textit{Outer score}, which differentiates the binary assignments to \textsc{kkc} and \textsc{uuc}, hides the actual efficiency of the model and makes the presented result dependent not on the actual quality of recognition, but rather on the scale of proportions between the \textsc{kkc} and \textsc{uuc} cardinalites in the test set.

In the presented research, we extend the two popularly applied measures of recognition quality and assess the methods with four metrics:
\begin{itemize}
     \item \textit{Inner score} -- a metric calculated only within known classes, represents the quality of closed-set classification. The ability to recognize objects of unknown classes does not affect its value.
     \item \textit{Outer score} -- a metric calculated for a binary classification task, assesses only the ability to recognize within \textsc{kkc} and \textsc{uuc}. Value is not dependent on the model's ability to recognize within known classes.
     \item \textit{Halfpoint score} -- a modified \textit{Inner score} metric that considers False Unknowns (instances considered unknown by the classifier despite belonging to the \textsc{kkc}) in the assessment.
     \item \textit{Overall score} -- a metric that treats the \textsc{uuc} class as a single additional class, equivalent to \textsc{kkc}.
\end{itemize}

In order to calculate the considered scoring functions, specific classes are taken into account, depending on the determined metric. In the case of \textit{Inner score}, only disjoint classes belonging to \textsc{kkc} will be considered. With \textit{Outer score}, a binary problem will be scored. All classes belonging to \textsc{kkc} will be in one category (positive), and all classes belonging to \textsc{uuc} will be in a second category (negative). The \textit{Halfpoint score} considers both \textsc{kkc} and \textsc{uuc} disjoint classes; however, the labels do not contain \textsc{uuc} cases, meaning the algorithm will only be penalized for incorrect predictions towards \textsc{uuc}. The most general metric \textit{Overall score} treats all \textsc{uuc} classes as one additional category, equal to \textsc{kkc}. 

\begin{figure}[!htb]
    \centering
    \input{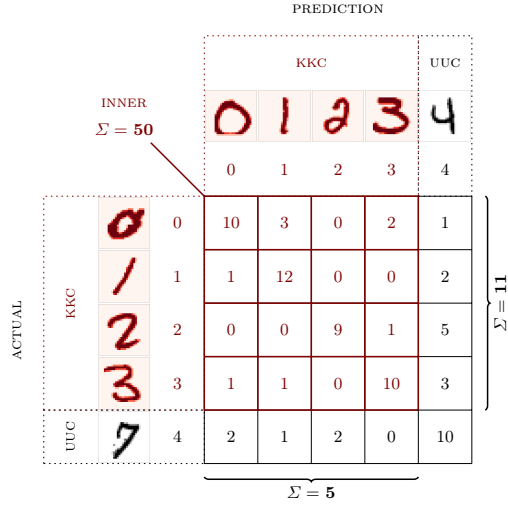}
    \caption{The general confusion matrix divided into specific regions subject to individual metrics}
    \label{fig:matrix1}
\end{figure}

A presentation of an exemplary confusion matrix based on which four measures are calculated is given in Figure~\ref{fig:matrix1}. The matrix considers the \textsc{uuc} class (in the last row and last column) as a separate \textsc{kkc} equivalent class. 

\begin{figure}[!htb]
    \centering
    \begin{center}
\resizebox{.55\textwidth}{!}{\begin{tikzpicture}
	
\begin{scope}[shift={(-7,-7)}]

\begin{scope}[shift={(-.5,0)}]
\draw[step=.5cm, red!50!black, thick] (3.99,0) grid (6,2);
\node[red!50!black, font=\tiny, align=left, text width=2cm] at (5,2.25) {\textsc{inner}};
\begin{scope}[shift={(4.25,1.75)}, font=\tiny]
	\node[red!50!black] at (0,0) {11};
	\node[red!50!black] at (.5,0) {3};
	\node[red!50!black] at (1,0) {0};
	\node[red!50!black] at (1.5,0) {2};
	
	\node[red!50!black] at (0,-.5) {3};
	\node[red!50!black] at (.5,-.5) {12};
	\node[red!50!black] at (1,-.5) {0};
	\node[red!50!black] at (1.5,-.5) {0};

	\node[red!50!black] at (0,-1) {0};
	\node[red!50!black] at (.5,-1) {0};
	\node[red!50!black] at (1,-1) {13};
	\node[red!50!black] at (1.5,-1) {2};

	\node[red!50!black] at (0,-1.5) {4};
	\node[red!50!black] at (.5,-1.5) {1};
	\node[red!50!black] at (1,-1.5) {0};
	\node[red!50!black] at (1.5,-1.5) {10};
\end{scope}
\end{scope}

\begin{scope}[shift={(-3.5,-3)}]

\draw[fill=black!10!white] (7,-.5) rectangle (9.5,0);
\draw[step=.5cm] (6.99,-.5) grid (9.5,2);

\draw[red!50!black, step=.5cm, thick] (6.99,0) grid (9,2);
\node[font=\tiny, align=left, text width=2cm] at (8,2.25) {\textsc{half-point}};

\begin{scope}[shift={(7.25,1.75)}, font=\tiny]
	\node[red!50!black] at (0,0) {10};
	\node[red!50!black] at (.5,0) {3};
	\node[red!50!black] at (1,0) {0};
	\node[red!50!black] at (1.5,0) {2};
	\node[] at (2,0) {1};
	
	\node[red!50!black] at (0,-.5) {1};
	\node[red!50!black] at (.5,-.5) {12};
	\node[red!50!black] at (1,-.5) {0};
	\node[red!50!black] at (1.5,-.5) {0};
	\node[] at (2,-.5) {2};

	\node[red!50!black] at (0,-1) {0};
	\node[red!50!black] at (.5,-1) {0};
	\node[red!50!black] at (1,-1) {9};
	\node[red!50!black] at (1.5,-1) {1};
	\node[] at (2,-1) {5};

	\node[red!50!black] at (0,-1.5) {1};
	\node[red!50!black] at (.5,-1.5) {1};
	\node[red!50!black] at (1,-1.5) {0};
	\node[red!50!black] at (1.5,-1.5) {10};
	\node[] at (2,-1.5) {3};

	\node[] at (0,-2) {\emph{0}};
	\node[] at (.5,-2) {\emph{0}};
	\node[] at (1,-2) {\emph{0}};
	\node[] at (1.5,-2) {\emph{0}};
	\node[] at (2,-2) {\emph{0}};
\end{scope}
\end{scope}

\begin{scope}[shift={(2.5,3.5)}]

\draw[step=.5cm] (3.99,-1.5) grid (5,-2.5);
\draw[red!50!black, step=.5cm, thick] (3.99,-1.5) grid (4.5,-2);
\node[font=\tiny, align=left, text width=2cm] at (5,-1.25) {\textsc{outer}};
\begin{scope}[shift={(4.25,-1.75)}, font=\tiny]
	\node[red!50!black] at (0,0) {\bfseries 50};
	\node[] at (.5,0) {\bfseries11};
	
	\node[] at (0,-.5) {\bfseries 5};
	\node[] at (.5,-.5) {10};
\end{scope}

\end{scope}

\begin{scope}[shift={(-.5,.5)}]

\draw[step=.5cm] (6.99,-4) grid (9.5,-1.5);
\draw[red!50!black, step=.5cm, thick] (6.99,-3.5) grid (9,-1.5);

\node[font=\tiny, align=left, text width=2cm] at (8,-1.25) {\textsc{overall}};
\begin{scope}[shift={(7.25,-1.75)}, font=\tiny]
	\node[red!50!black] at (0,0) {10};
	\node[red!50!black] at (.5,0) {3};
	\node[red!50!black] at (1,0) {0};
	\node[red!50!black] at (1.5,0) {2};
	\node[] at (2,0) {1};
	
	\node[red!50!black] at (0,-.5) {1};
	\node[red!50!black] at (.5,-.5) {12};
	\node[red!50!black] at (1,-.5) {0};
	\node[red!50!black] at (1.5,-.5) {0};
	\node[] at (2,-.5) {2};

	\node[red!50!black] at (0,-1) {0};
	\node[red!50!black] at (.5,-1) {0};
	\node[red!50!black] at (1,-1) {9};
	\node[red!50!black] at (1.5,-1) {1};
	\node[] at (2,-1) {5};

	\node[red!50!black] at (0,-1.5) {1};
	\node[red!50!black] at (.5,-1.5) {1};
	\node[red!50!black] at (1,-1.5) {0};
	\node[red!50!black] at (1.5,-1.5) {10};
	\node[] at (2,-1.5) {3};

	\node[] at (0,-2) {2};
	\node[] at (.5,-2) {1};
	\node[] at (1,-2) {2};
	\node[] at (1.5,-2) {0};
	\node[] at (2,-2) {10};
\end{scope}
\end{scope}
\end{scope}
	
\end{tikzpicture}}
\end{center}
    \caption{Examples of derived confusion matrices based on the general matrix in Figure~\ref{fig:matrix1}}
    \label{fig:matrix2}
\end{figure}
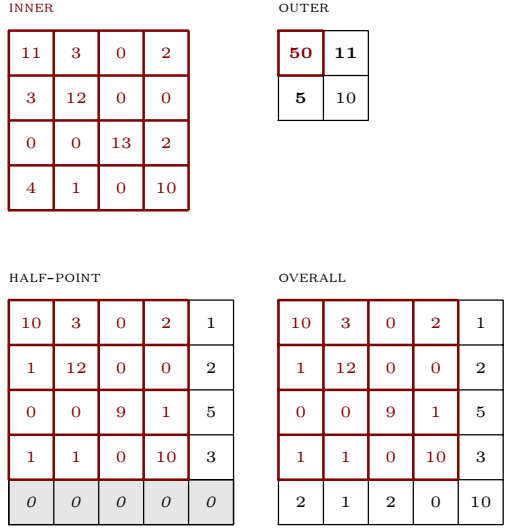

\textsc{uuc} samples will include instances of all classes marked as unknown. Based on the sample matrix in Figure~\ref{fig:matrix2}, modified confusion matrices are presented, which will be used to determine the four proposed measures. It is worth noting that in the case of the matrix for the \textit{Inner score}, the values for individual fields differ from those presented in the matrix for the \emph{Overall score}. The closed-set accuracy calculation requires that all \textsc{kkc} samples in the test set also return the class of the \textsc{kkc} set. Thus, instances that would be mistakenly recognized as \textsc{uuc} are assigned to the most likely category from the \textsc{kkc}. The matrix is flipped in the \textit{Outer score} for better presentation. 

In the final calculation, according to the definition of the \textsc{osr} problem, \textsc{kkc} instances will state a positive class and \textsc{uuc} instances a negative class \cite{scheirer2012toward}. In the matrix for \textit{Halfpoint score}, the last row contains zeros only, which indicates that the algorithm will be penalized only for False Unknown predictions in the last column.

\section{Exemplary experimental evaluation}

This section presents an example experiment comparing the operation of four simple methods. Its primary purpose, however, is to present all the components of the experimental protocol.

\subsection{Datasets}

Open Set Recognition considers \textsc{uuc} recognition, which indicates that no prior information or samples of unknown classes are available. However, for experimental analysis purposes only, it is necessary to sample \textsc{uuc} instances to simulate the emergence of new, unknown classes in end-uses of algorithms \cite{geng2020recent}. 

The evaluation used the \textit{Outlier} protocol -- where \textsc{kkc} instances are taken from one dataset and \textsc{uuc} from another external dataset that may contain samples from different domains \cite{bulusu2020anomalous}. In this experiment, \textsc{kkc} instances were sampled from the CIFAR10 dataset and \textsc{uuc} from the SVHN dataset. Both contain color images of size : $32\times 32$, describing ten different types of objects (CIFAR) or digits of house numbers seen from the street (SVHN). Another popularly used configuration of this protocol type, mentioned in the previous section, assumes sampling \textsc{kkc} from the MNIST dataset, describing handwritten digits and \textsc{uuc} from the Omniglot dataset describing handwritten letters from various alphabets \cite{yoshihashi2019classification}.

\begin{figure}[!htb]
    \centering
    \includegraphics[width=0.35\textwidth]{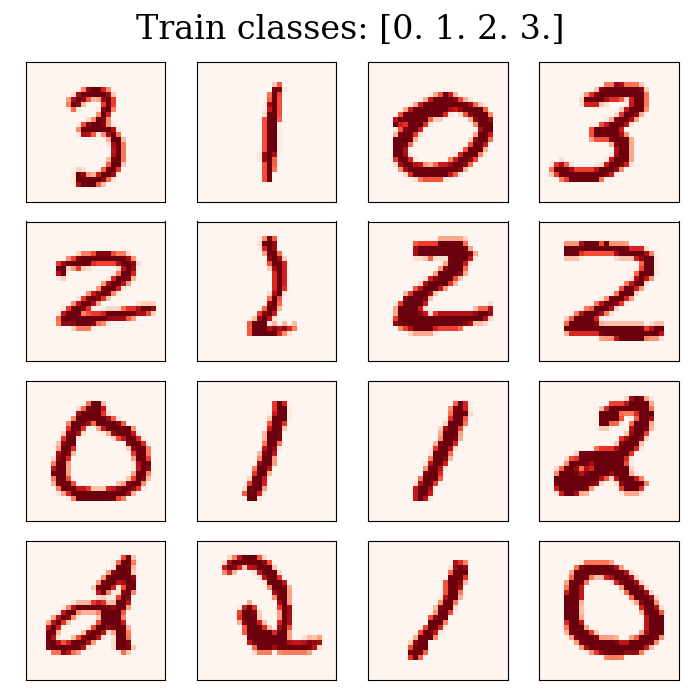}
    \includegraphics[width=0.35\textwidth]{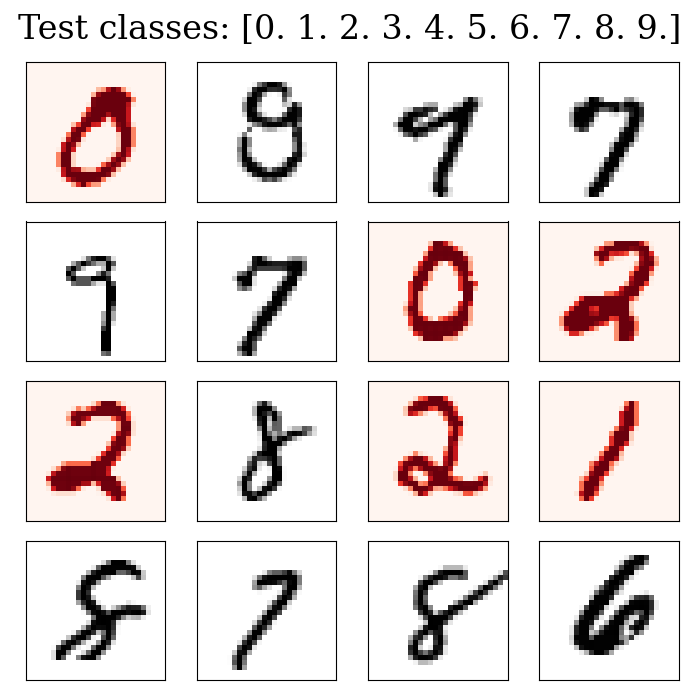}
    \caption{An example of randomly selected samples of Training and Testing sets, where red digits come from \textsc{kkc} and blue from \textsc{uuc}}
    \label{fig:train_test}
\end{figure}

In the second experimental protocol type -- \textit{Holdout} -- only part of the classes of a multiclass dataset is used to train the model, and a disjoint set of classes from the same dataset is considered as \textsc{uuc} and used only in the testing process. Figure \ref{fig:train_test} shows an example of data from the MNIST dataset using \textit{Holdout} protocol. The training set contains only data from \textsc{kkc}, but the testing set includes both samples from \textsc{kkc} and \textsc{uuc}.  

In the exemplary experiment, first the cardinalities of \textsc{kkc} and \textsc{uuc} were determined (and therefore the \textit{Openness} of data configuration). Then, within a given \textit{Openness}, specific known and unknown classes were randomly selected. The methods were evaluated on five different \textit{Openness} values, and, within a single class cardinality configuration, classes were assigned to \textsc{kkc}/\textsc{uuc} categories five times. Such an approach resulted in 25 different \textsc{osr} problems inferred from two 10-class datasets. Each dataset prepared in such a way was additionally divided into two folds to perform cross-validation. Instances of unknown classes were removed from the training sets and used only in the testing process. 

\subsection{Method configuration and evaluation measures}

In the presented experiment, four simple approaches were evaluated -- two discriminative and two generative methods. The short description of evaluated solutions is as follows:

\begin{itemize}
    \item \textbf{Thresholded Softmax} -- a baseline discriminative method using probability thresholding at a final layer of Neural Network.
    \item \textbf{Openmax} \cite{bendale2016towards} -- discriminative method using Weibull Distributions to approximate \textsc{kkc} class occurrence in the embedding's space.
    \item \textbf{Noise Softmax} -- generative method, which creates artificial \textsc{uuc} samples by generating images containing random noise in each pixel.
    \item \textbf{Overlay Softmax} -- generative method, creating artificial \textsc{uuc} samples by averaging two randomly selected \textsc{kkc} instances.
\end{itemize}

A selected network architecture was uniform for all evaluated methods. The architecture was similar to that of the research on the OSRCI method \cite{neal2018open}, containing two convolutional layers, two max-pooling layers, and a single dropout layer, along with ReLU activation functions and two fully connected layers as the final ones. The Cross-Entropy loss function was selected for all methods and the Stochastic Gradient Descent optimizer with a default approach to regularization. The networks were trained with batches of size 64 in 128 iterations, and the quality was measured after each epoch.

As mentioned in Section 3, class imbalance becomes a significant factor in the \textsc{osr} evaluation, especially considering the \textsc{kkc}/\textsc{uuc} category dichotomy. The methods were evaluated using four measures: \textit{Inner}, \textit{Outer}, \textit{Halfpoint}, and \textit{Overall} scores, using a Balanced Accuracy Score as a base measure.

\subsection{Analysis of obtained results}

The results of described experiment are presented in Table~\ref{tab:results}. It presents four metrics (\textit{Inner}, \textit{Outer}, \textit{Halfpoint}, \textit{Overall}) after the last epoch of training. The columns contain information about the \textit{Openness} of the set and the results for all four methods. The presented results are the average from 5 repetitions for a given \textit{Openness}, and their standard deviation is presented in brackets.

\begin{table}[]
    \centering
    \caption{Results of performed experiment in four metrics, averaged for examined Openness of datasets}
    
\scriptsize
\setlength{\tabcolsep}{5pt}
\renewcommand{\arraystretch}{1.5}

\begin{tabular}{|l|l|llll|}
\hline
                           & \textbf{Openness} & \textbf{Thresholded}      & \textbf{Openmax}       & \textbf{Noise}  & \textbf{Overlay} \\
                           & & \textbf{Softmax}      &        & \textbf{Softmax} & \textbf{Softmax} \\ \hline

\multirow{5}{*}{\textit{Inner}}     & 0.142    & 0.515 (0.025) & \textbf{0.518} (0.026) & 0.507 (0.027) & 0.510 (0.030)  \\
                           & 0.147    & 0.716 (0.031) & \textbf{0.718} (0.035) & 0.691 (0.031) & 0.692 (0.034)  \\
                           & 0.202    & 0.529 (0.023) & \textbf{0.537} (0.019) & 0.527 (0.017) & 0.525 (0.027)  \\
                           & 0.22     & 0.529 (0.022) & \textbf{0.536} (0.027) & 0.520 (0.021) & 0.526 (0.026)  \\
                           & 0.423    & 0.852 (0.065) & \textbf{0.853} (0.064) & 0.819 (0.094) & 0.821 (0.084)  \\\hline
\multirow{5}{*}{\textit{Outer}}     & 0.142    & 0.502 (0.004) & 0.500 (0.000) & 0.500 (0.000) & \textbf{0.752} (0.017)  \\
                           & 0.147    & 0.521 (0.040) & 0.534 (0.035) & 0.490 (0.029) & \textbf{0.676} (0.036)  \\
                           & 0.202    & 0.502 (0.003) & 0.500 (0.000) & 0.499 (0.000) & \textbf{0.747} (0.016)  \\
                           & 0.22     & 0.501 (0.004) & 0.500 (0.000) & 0.499 (0.000) & \textbf{0.762} (0.021)  \\
                           & 0.423    & 0.635 (0.044) & 0.500 (0.000) & 0.600 (0.037) & \textbf{0.680} (0.076)  \\\hline
\multirow{5}{*}{\textit{Halfpoint}} & 0.142    & \textbf{0.446} (0.023) & 0.453 (0.023) & 0.444 (0.023) & 0.405 (0.033)  \\
                           & 0.147    & 0.320 (0.028) & 0.284 (0.040) & \textbf{0.448} (0.038) & 0.381 (0.039)  \\
                           & 0.202    & 0.458 (0.022) & \textbf{0.470} (0.017) & 0.461 (0.015) & 0.412 (0.021)  \\
                           & 0.22     & 0.459 (0.020) & \textbf{0.469} (0.023) & 0.455 (0.018) & 0.422 (0.028)  \\
                           & 0.423    & 0.288 (0.130) & 0.000 (0.000) & \textbf{0.408} (0.109) & 0.274 (0.113)  \\\hline
\multirow{5}{*}{\textit{Overall}}   & 0.142    & 0.449 (0.022) & 0.453 (0.023) & 0.444 (0.023) & \textbf{0.485} (0.027)  \\
                           & 0.147    & 0.436 (0.025) & 0.417 (0.031) & 0.499 (0.037) & \textbf{0.532} (0.033)  \\
                           & 0.202    & 0.461 (0.020) & 0.470 (0.017) & 0.461 (0.015) & \textbf{0.491} (0.018)  \\
                           & 0.22     & 0.461 (0.019) & 0.469 (0.023) & 0.455 (0.018) & \textbf{0.502} (0.021)  \\
                           & 0.423    & 0.560 (0.071) & 0.333 (0.000) & \textbf{0.585} (0.075) & 0.579 (0.111) \\\hline

\end{tabular}

    \label{tab:results}
\end{table}

In the case of \textit{Inner score}, the strategy of discriminative methods (Thresholded Softmax and Openmax) is consistent, therefore the results for them are similar. The noticeably lower quality of recognition within known classes by generative methods may result from the need to recognize an additional class (artificially created unknown class), which makes it difficult to optimize the weights for a smaller pool of classes within which the evaluation takes place.
In the case of \textit{Outer score} and this problem, the advantage of the Overlay Softmax method is clearly visible. However, this method did not prove to be the best for the \textit{Halfpoint} score in any of the tested \textit{Opennesses}. \textit{Halfpoint} score extends \textit{Inner score} by taking into account False Negatives -- incorrect assignment of known objects to an unknown category. It is also noticeable that the Thresholded Softmax and Openmax methods have lost the advantage visible in the \textit{Inner score} -- which may suggest a large number of false unknowns in these methods. An extreme case is the result of 0\% for Openmax at the highest \textit{Openness} -- this means that all objects were incorrectly recognized as unknown.
In the \textit{Overall score}, the best results are achieved by Overlay Softmax and Noise Softmax, which confirms that generative methods constitute a valuable pool of methods in the field.

\begin{figure}
    \centering
    \includegraphics[width=\textwidth]{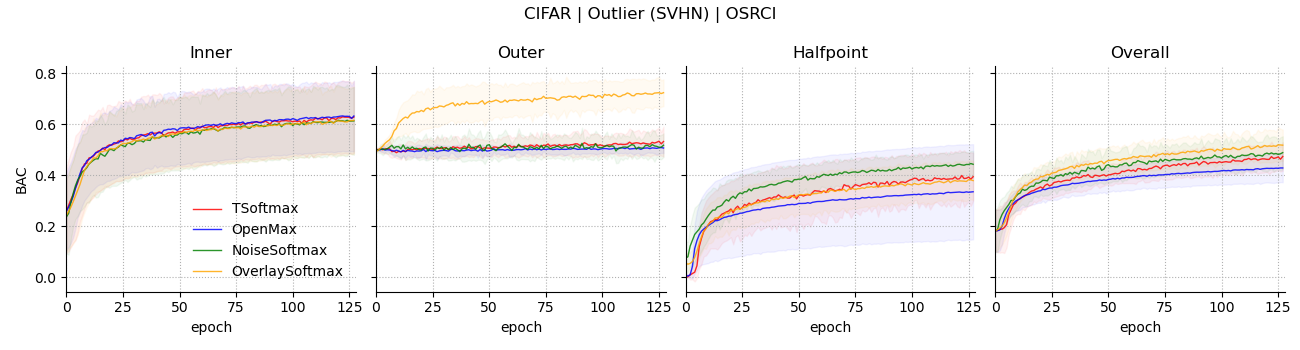}
    \caption{Metrics' results over a training process, averaged in all configurations}
    \label{fig:example}
\end{figure}

Additionally, the learning process over all epochs is shown in Figure~\ref{fig:example}. The results in the graph are averaged over all 25 configurations, consisting of 5 \textit{Openness} and 5 repetitions of assigning classes to categories. The standard deviation is also shown with decreased opacity.

The learning process shows a consistent increase in quality across all methods. In \textit{Inner score}, learning advances similar for all methods. Greater differences are visible in the remaining metrics. In the \textit{Outer score} there is a large advantage for the Overlay Softmax method, in the \textit{Halfpoint score} for the Noise Softmax method, and in the \textit{Overall score} there is a slight advantage of the generative methods over Thresholded Softmax and Openmax.

It is worth to keep in mind that such an evaluation was performed for a specific pair of datasets, specific experimental protocol type (\textit{Outlier}) and a single network architecture. To draw conclusions about the general method advantage over reference approaches, those variables should as well be taken into account during evaluation.

\section{Conclusions}

The work discusses the topic of non-trivial experimental evaluation in the Open Set environment. The discussion focuses on the context of uneven class distribution, particularly in the case of dichotomies between known and unknown objects, in standard experimental protocols in the field.

As a result of the problem analysis we present a set of guidelines aimed to be valuable for researchers for evaluation in the Open Set:

\begin{itemize}
     \item Algorithms should be evaluated for multiple problem configurations, offering a variability of \textit{Openness} values, depending on the size of \textsc{kkc} and \textsc{uuc} classes.
     \item Within each \textit{Openness}, multiple repetitions should be considered, in which specific classes will be assigned to \textsc{kkc} and \textsc{uuc} categories.
     \item Instead of the \textit{Inner score} measure used so far, it is worth using the proposed \textit{Halfpoint score}, which considers False Unknowns.
     \item It is worth using measures to aggregate recognition ability in both closed-set and open-set, such as the presented \textit{Halfpoint score} and \textit{Overall score}, in addition to \textit{Outer score}.
     \item The problem imbalance ratio in terms of \textsc{kkc}/\textsc{uuc} dichotomy should be monitored and considered when assessing the ability to recognize unknown objects.
     \item Quality assessment using the F-score and Accuracy metrics should be avoided due to the strong asymmetry of the metric and the incapacity to compare the results for different test sets.
     \item Optimal method hyperparameter selection should depend solely on the number of \textsc{kkc} and their characteristics. The \textit{Openness} values, often used to determine the optimal hyperparameters, are developed only for descriptions of sets for evaluation and will not be known in final applications.
\end{itemize}

\section*{Acknowledgement}
This work was supported by the statutory funds of the Department of Systems and Computer Networks, Faculty of Information and Communication Technology, Wrocław University of Science and Technology.

\bibliographystyle{splncs04} 
\bibliography{biblio}

\end{document}